\documentclass[conference]{IEEEtran}
\IEEEoverridecommandlockouts
\usepackage{cite}
\usepackage{amsmath,amssymb,amsfonts}
\usepackage{graphicx}
\usepackage{textcomp}
\usepackage{xcolor}
\usepackage{bm}
\usepackage{amsmath}
\usepackage{amssymb}
\usepackage{amsthm}
\usepackage{mathrsfs}
\usepackage{enumerate}
\usepackage{multirow}
\usepackage{color}
\usepackage{threeparttable}
\usepackage{subfigure}

\usepackage[square, comma, sort&compress, numbers]{natbib}
\usepackage[pagebackref=false,breaklinks=true,letterpaper=true,colorlinks,bookmarks=false]{hyperref}
\usepackage{times}
\usepackage{breakurl}
\usepackage{array}
\usepackage{verbatim}
\usepackage{algorithm}
\usepackage{bbding}
\usepackage{algpseudocode}
\usepackage{lettrine}

\usepackage{booktabs}
\usepackage{multirow}
\usepackage{colortbl} 

\definecolor{dc1}{RGB}{0,0,255} 
\definecolor{dc2}{RGB}{0,165,255}
\definecolor{dc3}{RGB}{0,255,255}
\definecolor{dc4}{RGB}{0,255,128}
\definecolor{dc5}{RGB}{0,255,0}
\definecolor{dc6}{RGB}{128,255,0}
\definecolor{dc7}{RGB}{255,255,0}
\definecolor{dc8}{RGB}{255,128,0}
\definecolor{dc9}{RGB}{255,0,0}
\definecolor{dc10}{RGB}{226,43,138}
\definecolor{dc11}{RGB}{230,216,173}
\definecolor{dc12}{RGB}{128,128,240}
\definecolor{dc13}{RGB}{180,105,255}
\definecolor{dc14}{RGB}{255,0,255}
\definecolor{dc15}{RGB}{212,0,148}
\definecolor{dc16}{RGB}{214,112,218}
\definecolor{dc17}{RGB}{147,20,255}
\definecolor{dc18}{RGB}{70,100,255}
\definecolor{dc19}{RGB}{0,140,255}
\definecolor{dc20}{RGB}{140,230,240}

\def\BibTeX{{\rm B\kern-.05em{\sc i\kern-.025em b}\kern-.08em
    T\kern-.1667em\lower.7ex\hbox{E}\kern-.125emX}}
\begin{document}

\title{Fab-ME: A Vision State-Space and Attention-Enhanced Framework for Fabric Defect Detection\\ 
\thanks{*Corresponding author. 

This work was supported by the Shandong Youth University of Political Science Doctoral Research Startup Fund (XXPY23036).}
}

\author{\IEEEauthorblockN{1\textsuperscript{st} Shuai Wang}
\IEEEauthorblockA{\textit{School of Information Engineering} \\
\textit{Shandong Youth University of Political Science}\\
Jinan, China \\
email:230013@sdyu.edu.cn}
\and
\IEEEauthorblockN{2\textsuperscript{nd} Huiyan Kong}
\IEEEauthorblockA{\textit{School of Information Engineering} \\
\textit{Shandong Youth University of Political Science}\\
Jinan, China \\
email:khy2345@163.com}
\and
\IEEEauthorblockN{3\textsuperscript{rd} Baotian Li}
\IEEEauthorblockA{\textit{School of Information Engineering} \\
\textit{Shandong Youth University of Political Science}\\
Jinan, China \\
email:sdyu\_lbt@163.com}
\and
\IEEEauthorblockN{4\textsuperscript{th} Fa Zheng* }
\IEEEauthorblockA{\textit{School of Information Engineering} \\
\textit{Shandong Youth University of Political Science}\\
Jinan, China \\
email:ffzheng0410@163.com}
}

\maketitle

\begin{abstract}
Effective defect detection is critical for ensuring the quality, functionality, and economic value of textile products. However, existing methods face challenges in achieving high accuracy, real-time performance, and efficient global information extraction. To address these issues, we propose Fab-ME, an advanced framework based on YOLOv8s, specifically designed for the accurate detection of 20 fabric defect types. Our contributions include the introduction of the cross-stage partial bottleneck with two convolutions (C2F) vision state-space (C2F-VMamba) module, which integrates visual state-space (VSS) blocks into the YOLOv8s feature fusion network neck, enhancing the capture of intricate details and global context while maintaining high processing speeds. Additionally, we incorporate an enhanced multi-scale channel attention (EMCA) module into the final layer of the feature extraction network, significantly improving sensitivity to small targets. Experimental results on the Tianchi fabric defect detection dataset demonstrate that Fab-ME achieves a 3.5\% improvement in mAP@0.5 compared to the original YOLOv8s, validating its effectiveness for precise and efficient fabric defect detection.

\end{abstract}


\section{Introduction}\label{sec:intro}

The textile industry plays a critical role in the global economy, yet ensuring fabric quality remains a significant challenge in the era of intelligent manufacturing. Accurate defect detection~\cite{kahramanDeepLearningbasedFabric2023, yan2024enhancing, tang2024fourier} is vital to maintain both aesthetic and functional standards, driving the development of high-precision, real-time inspection systems. Defects such as misalignments and structural irregularities not only compromise product quality but also impact manufacturers' financial stability and brand reputation.

Deep learning has emerged as a transformative approach to fabric defect detection, enabling the automatic learning of complex image features and significantly improving the recognition of intricate textures and diverse defect types. Convolutional neural networks (CNNs) have outperformed traditional methods in tasks such as detection~\cite{jiaFabricDefectDetection2022,weng2024enhancing,li2024lr}, segmentation~\cite{liuFPDeeplabSegmentationModel2024}, and generation~\cite{shen2023advancing, shen2024boosting, shen2024imagdressing, shen2024imagpose}. Beyond CNNs, advanced models including autoencoders~\cite{zhangMixedAttentionBased2023}, knowledge distillation~\cite{zhangKnowledgeDistillationUnsupervised2024}, and memory banks~\cite{wuAutomatedFabricDefect2024} have introduced innovative perspectives.

\begin{figure*}[t]
    \vspace{-0.5cm}
    \centering
    \includegraphics[width=0.9\linewidth]{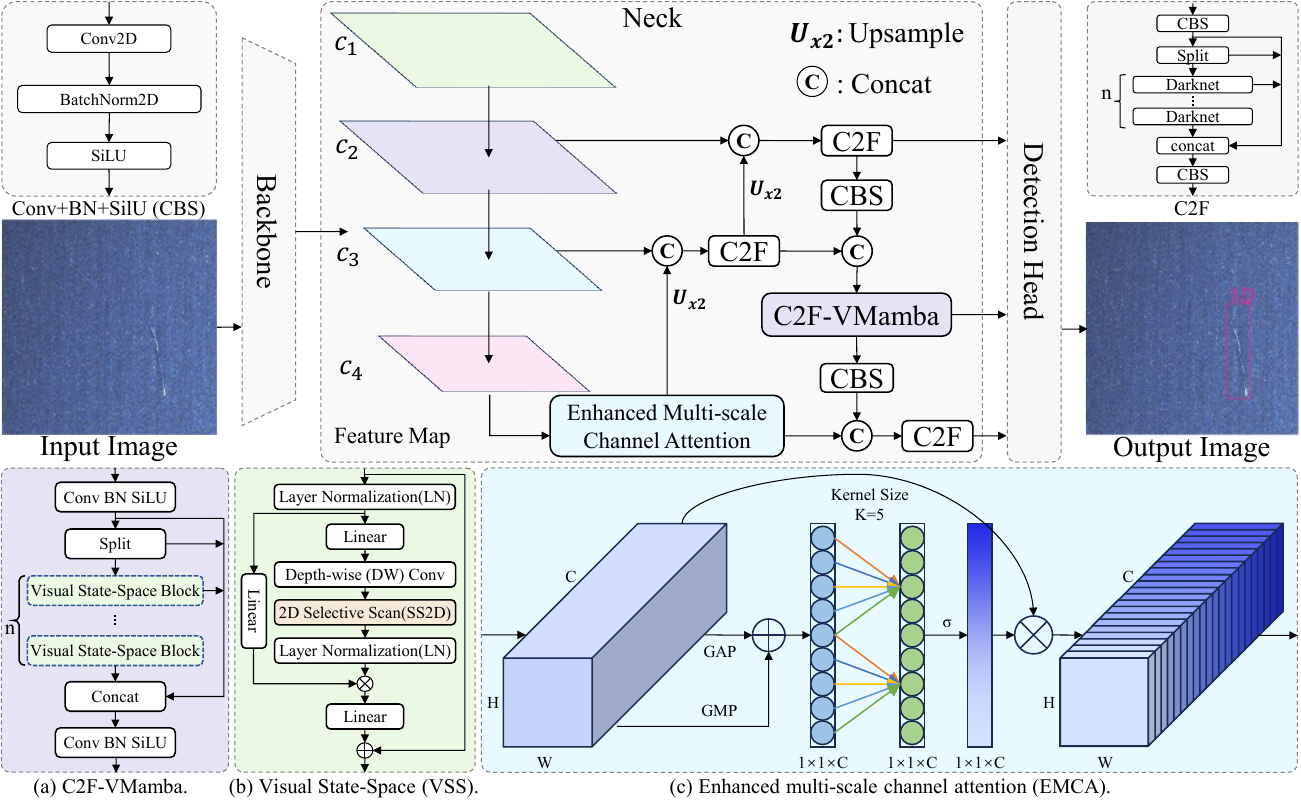}
    \vspace{-0.20cm}
\caption{The proposed Fab-ME framework. The feature maps generated at each stage of the backbone network are denoted as $c_1$ through $c_4$. C2F stands for "CSP Bottleneck with Two Convolutions. (a) C2F-VMamba. (b) Visual State-Space (VSS). (c) Enhanced multi-scale channel attention (EMCA).}
\label{fig:framework}
\vspace{-0.2cm}
\end{figure*}

In the domain of defect detection, object detection methods leveraging CNNs and transformers have gained prominence. Two-stage approaches, such as the R-CNN series~\cite{renFasterRCNNRealtime2016}, achieve high accuracy by refining classification and localization but are computationally expensive. In contrast, one-stage methods, such as YOLO~\cite{jocher2020ultralytics,meiResearchFabricDefect2024} and SSD~\cite{zhaoFabricSurfaceDefect2022}, combine localization and classification into a single step, making them ideal for real-time applications. Recent innovations, including attention mechanisms~\cite{qiao2022novel,weng2023novel}, large receptive field modules~\cite{zhangMixedAttentionBased2023}, and feature fusion modules~\cite{liuPRClightYOLOEfficient2024, liu2024feature}, have significantly enhanced detection accuracy.
However, existing models often overlook multi-scale information, which is critical for detecting defects of varying sizes and complexities. Furthermore, many methods rely on either global or local features, limiting their ability to capture the intricate details necessary for precise defect detection. These limitations highlight the need for an integrated approach that effectively combines global and local features.

The Mamba framework~\cite{zhuVisionMambaEfficient2024} has recently demonstrated strong potential in addressing these challenges. Vision Mamba~\cite{zhuVisionMambaEfficient2024} introduces bidirectional Mamba blocks to enhance visual data representation, while MambaAD~\cite{heMambaadExploringState2024} incorporates multi-scale state-space modules for anomaly detection. These advancements provide a foundation for addressing multi-scale and context-aware defect detection challenges.

In this paper, we present Fab-ME, an enhanced framework based on YOLOv8s, for detecting 20 types of fabric defects. To address challenges in detecting large targets, we integrate visual state-space (VSS) blocks~\cite{liuVMambaVisualState2024} into the cross-stage partial bottleneck with two convolutions (C2F) module, forming the C2F-VMamba module. This integration improves global context understanding and detail preservation while maintaining real-time performance. Additionally, an enhanced multi-scale channel attention (EMCA) module is incorporated into the feature extraction network, boosting sensitivity to small targets. Experimental results on the Tianchi fabric defect detection dataset show a 3.5\% improvement in mAP@0.5 over YOLOv8s, with real-time processing capabilities.
The main contributions of this work are summarized as follows:
\begin{itemize}
    \item We propose the C2F-VMamba module, integrating visual state-space (VSS) blocks into YOLOv8s, enabling efficient global context capture and intricate detail extraction.
    \item We design an enhanced multi-scale channel attention (EMCA) module, improving multi-scale feature extraction and sensitivity to small targets.
    \item The Fab-ME framework achieves a 3.5\% mAP@0.5 improvement on the Tianchi dataset while maintaining real-time performance.
\end{itemize}

\section{Proposed Method}\label{sec:method} 
\subsection{Overview}
This paper introduces an improved YOLOv8 algorithm ~\cite{sohanReviewYolov8Its2024}, featuring two primary enhancements: the integration of the EMCA module after the spatial pyramid pooling fast (SPPF) ~\cite{zhu2021tph} block within the YOLOv8 backbone, and the substitution of the third C2F block in the neck with the C2F-VMamba module. The proposed architecture, Fab-ME, is illustrated in Fig. ~\ref{fig:framework}.
In the feature fusion module neck, the light blue box denotes the EMCA module, which is incorporated to enhance the extraction of discriminative features across multiple scales. The light purple box labeled "C2F-VMamba" in the neck section denotes replacing the original C2F block with the C2F-VMamba module, enhancing the model's feature transformation capabilities.

\subsection{C2F-VMamba}  

The C2F module is an essential element for feature extraction in YOLOv8s, leveraging CSP network principles and residual structures. The architecture comprises multiple convolutional layers and bottleneck modules designed to capture and utilize rich gradient flow information effectively. To enhance global feature extraction in the neck, we propose the C2F-VMamba module. This module substitutes specific convolutional layers in the C2F module with VSS blocks~\cite{liuVMambaVisualState2024}. The C2F-VMamba module maintains the lightweight architecture and efficient gradient flow information extraction of the C2F module. It also incorporates the global receptive field and linear complexity benefits of VSS blocks, enhancing the detection of fabric defect features.

The C2F-VMamba module (see Fig. \ref{fig:framework} (a)) operates as follows:

The input feature map \( X \) is processed through an initial convolutional layer, resulting in \(\text{Conv}(X)\). This output is then split into two parts, \( X_1 \) and \( X_2 \).
$$
   X_1, X_2 = \text{Split}(\text{Conv}(X)).
$$

The segment \( X_2 \) is passed through a Visual State-Space (VSS) module to produce the feature map \( Y_2 \).
$$
   Y_2 = \text{VSS}(X_2).
$$

The output \( Y_2 \) undergoes further processing through \( n-1 \) additional VSS modules, yielding \( Y_2' \).
$$
   Y_2' = \underbrace{\text{VSS}(\text{VSS}(\ldots \text{VSS}(Y_2) \ldots))}_{n-1 \text{ times}}.
$$

The features \( X_1 \), \(\text{Conv}(X)\), \( Y_2 \), and \( Y_2' \) are concatenated along the channel dimension. The concatenated features are then passed through a final convolutional layer to produce the output feature map.
$$
   \text{Output} = \text{Conv}(\text{Concat}(X_1, \text{Conv}(X), Y_2, Y_2')).
$$

Within the VSS blocks (refer to Fig. \ref{fig:framework} (b)), the feature map is bifurcated into two distinct paths. One path undergoes depthwise convolution, while the other is processed through 2D selective scanning (SS2D)~\cite{liuVMambaVisualState2024} operations. The outputs from both paths are subsequently merged.

Integrating the C2F module from YOLOv8s with the VSS blocks from the VMamba model results in the C2F-VMamba module, significantly improving fabric defect detection performance.
This module preserves the lightweight architecture and gradient flow information extraction of the C2F module while incorporating the global receptive field and linear complexity benefits of VSS blocks. This integration provides an efficient solution for fabric defect detection.

\begin{table}[t]
    \vspace{-0.50cm}
    \centering
    \caption{Distribution of defects in the dataset. "Img" denotes the number of images, and "Ann" indicates the number of annotated bounding boxes for each defect category.}\label{table:samples}
    \begin{tabular}{@{}ccp{2.25cm}cccc@{}}
    \toprule
    \multirow{2}{*}{ID} &\multirow{2}{*}{Color}&\multirow{2}{*}{Name} 
    & \multicolumn{2}{c}{Train   Set} & \multicolumn{2}{c}{Val Set} \\ \cmidrule(l){4-7} 
        & & & Img       & Ann      & Img     & Ann    \\ \midrule
    1  &\cellcolor{dc1}& holes                & 409   & 486   & 93   & 114 \\
    2  &\cellcolor{dc2}& water stains, etc.   & 609   & 768   & 158  & 205 \\
    3  &\cellcolor{dc3}& three-yarn defects   & 1550 & 1698 & 267  & 276 \\
    4  &\cellcolor{dc4}& knots                & 2561 & 3039 & 465  & 551 \\
    5  &\cellcolor{dc5}& pattern skips        & 640   & 682   & 153  & 154 \\
    6  &\cellcolor{dc6}& hundred-leg defects  & 804   & 875   & 130  & 140 \\
    7  &\cellcolor{dc7}& neps                 & 282   & 305   & 57   & 66  \\
    8  &\cellcolor{dc8}& thick ends           & 452   & 463   & 85   & 85  \\
    9  &\cellcolor{dc9}& loose ends           & 827   & 870   & 152  & 165 \\
    10 &\cellcolor{dc10}& broken ends          & 590   & 610   & 124  & 130 \\
    11 &\cellcolor{dc11}& sagging ends         & 333   & 347   & 50   & 54  \\
    12 &\cellcolor{dc12}& thick fibers         & 1293 & 1605 & 219  & 269 \\
    13 &\cellcolor{dc13}& weft shrinkage       & 861   & 1034 & 150  & 191 \\
    14 &\cellcolor{dc14}& sizing spots         & 1398 & 1462 & 329  & 344 \\
    15 &\cellcolor{dc15}& warp knots           & 451   & 758   & 82   & 141 \\
    16 &\cellcolor{dc16}& star skips, etc.     & 621   & 625   & 129  & 129 \\
    17 &\cellcolor{dc17}& broken spandex       & 589   & 869   & 115  & 139 \\
    18 &\cellcolor{dc18}& color shading, etc.  & 1972 & 2000 & 390  & 394 \\
    19 &\cellcolor{dc19}& abrasion marks, etc. & 1210 & 1252 & 220  & 238 \\
    20 &\cellcolor{dc20}& dead folds, etc.     & 789   & 874   & 169  & 174 \\\bottomrule
    \end{tabular}
    \vspace{-0.20cm}
    \end{table}

\begin{figure}[t]
    \vspace{-0.20cm}
    \centering
    \includegraphics[width=0.95\linewidth]{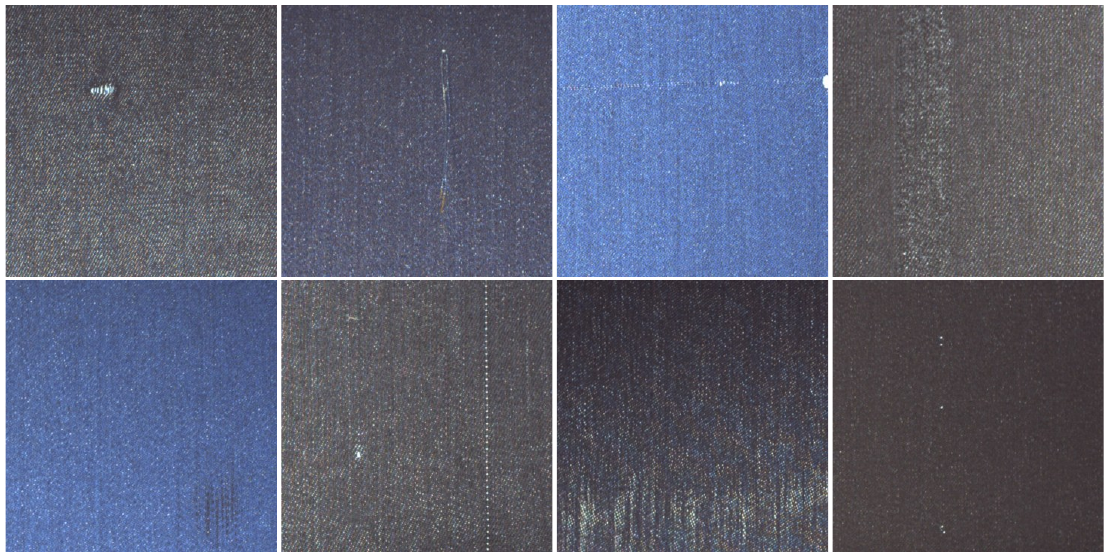}
    \vspace{-0.20cm}
\caption{Sample display of original fabric defect images from the Tianchi fabric dataset}
\label{fig:sample_image}
\vspace{-0.20cm}
\end{figure}

\subsection{Enhanced Multi-scale Channel Attention }  
Building upon the Efficient Channel Attention (ECA) module~\cite{wang2020eca}, this paper introduces the EMCA module. EMCA enhances multi-scale feature extraction while ensuring computational efficiency and maintaining a lightweight architecture. The ECA module efficiently captures channel-wise dependencies, balancing performance with complexity. However, it does not incorporate multi-scale integration. To address this limitation, EMCA integrates multi-scale contextual information, enriching feature representation and improving the model's capability to manage varying scales.

Given an input feature map \( \mathbf{F} \in \mathbb{R}^{H \times W \times C} \), the EMCA module computes attention as follows:

Compute the attention weights using combined channel descriptors:
$$
   \mathbf{a} = \sigma\left(\text{Conv1D}(\text{GAP}(\mathbf{F}) + \text{GMP}(\mathbf{F}), k)\right),
$$
where \( \sigma \) denotes the sigmoid activation function, and \( \mathbf{a} \in \mathbb{R}^{C} \) represents the attention weights for each channel.

Recalibrate the input feature map using the computed attention weights:
$$
   \mathbf{F'}(i, j, c) = \mathbf{a}(c) \cdot \mathbf{F}(i, j, c),
$$
for each channel \( c \), resulting in the output feature map \( \mathbf{F'} \).

\section{Experiment and Analysis}\label{sec:exp} 
To demonstrate the effectiveness of the proposed Fab-ME framework, we compare it against several state-of-the-art fabric defect detection methods using the large-scale Tianchi fabric dataset.

\subsection{Datasets}
\textbf{\emph{Tianchi fabric dataset}}
The dataset comprises 5,913 defect images and 3,663 non-defect images, with 9,523 annotations across 20 defect categories. Each image measures $2446 \times 1000$ pixels, and annotations are provided in YOLO and COCO formats. Images were partitioned into training and validation sets at a 4:1 ratio. To optimize YOLO training, images were segmented into $640 \times 640$ sub-images, discarding non-defective sub-images. This resulted in 17,152 training images and 3,360 validation images. Table~\ref{table:samples} summarizes image and annotation counts for each defect category.
The 20 defect categories include: holes; stains (water, oil, dirt); three-yarn defects; knots; pattern skips; hundred-leg defects; neps; thick ends; loose ends; broken ends; sagging ends; thick fibers; weft shrinkage; sizing spots; warp knots; star skips and skipped wefts; broken spandex; unevenness, waves, and color shading; abrasion, rolling, repair marks; and dead folds, cloud weaves, double wefts. Fig. \ref{fig:sample_image} presents examples of the 20 defect types. The dataset predominantly features long, narrow, and minor defects, which pose significant challenges for detection.

\subsection{Evaluation Metrics}  
In this paper, we adopt mean average precision at an IoU threshold of 0.5 (mAP@50) as the evaluation metric, which is a standard for object detection tasks. The mAP@50 is calculated by averaging the average precision (AP) of all categories, where AP represents the area under the precision-recall curve. Precision and recall are derived based on true positives, false positives, and false negatives, with an IoU threshold of 0.5 used to determine matches between predicted and ground truth bounding boxes. For the defect detection dataset, the mAP is computed as the mean of AP values across 20 categories.

\subsection{Implementation Details} 
The experiment was conducted on a server featuring eight NVIDIA RTX 4090 GPUs and an AMD EPYC 7551P CPU. Each training instance utilized eight CPU cores, 32 GB of RAM, and a single RTX 4090 GPU. The optimization process employed stochastic gradient descent (SGD) with a learning rate 0.005. 
The warmup period was set to 3 epochs, with a momentum of 0.937, a weight decay of \(1 \times 10^{-4}\), and a batch size 16. The training was terminated if performance metrics showed no improvement for 50 consecutive epochs, with the average training duration being 200 epochs.

\subsection{Comparison with State-of-the-art Methods} 
\begin{table}[t]
    \centering
    \caption{Comparison with state-of-the-art methods on Tianchi fabric dataset. A dash ("-") denotes unreported results in the respective paper.}\label{table:all_model_table}
    \begin{tabular}{{@{}ccccc@{}}}
        \toprule
    Methods            & mAP@0.5 (\%) & Param. (M)   & Year \\
     \midrule
    AMFF~\cite{zhaoAttentionBasedMultiscaleFeature2024} & 17.50 &   83.90    & 2024 \\
    C-RCNN~\cite{luFabricDefectDetection2023}           & 19.20 &   81.30    & 2023 \\
    Faster R-CNN~\cite{renFasterRCNNRealtime2016}       & 35.90 &   25.60    & 2016 \\
    Tood~\cite{feng2021tood}                            & 44.10 &   53.26    & 2021 \\
    FDDA~\cite{meiResearchFabricDefect2024}             & 49.80 &   -     & 2024 \\
    YOLOv5s~\cite{jocher2020ultralytics}                & 50.20 &   14.40     & 2020 \\
    PAMF~\cite{luAnchorFreeDefectDetector2023}          & 53.10 &   36.81    & 2023 \\
    YOLOX-CATD~\cite{wangFabricDefectDetection2023}     & 54.63 &   27.15    & 2023 \\
    AFAM~\cite{wangAdaptivelyFusedAttention2023}        & 56.70 &   69.63    & 2023 \\
    YOLO-TTD~\cite{yueResearchTinyTarget2022}           & 56.74 &   24.52    & 2022 \\
    \hline
    Baseline                                            & 57.40 &   11.10    & 2023 \\
    Fab-ME                                                & \textbf{59.40}     & \textbf{11.00}   &   Ours 
    \\ \bottomrule
\end{tabular}
\vspace{-0.20cm}
\end{table}

\begin{figure}[t]
    \vspace{-0.20cm}
    \centering
    \includegraphics[width=0.9\linewidth]{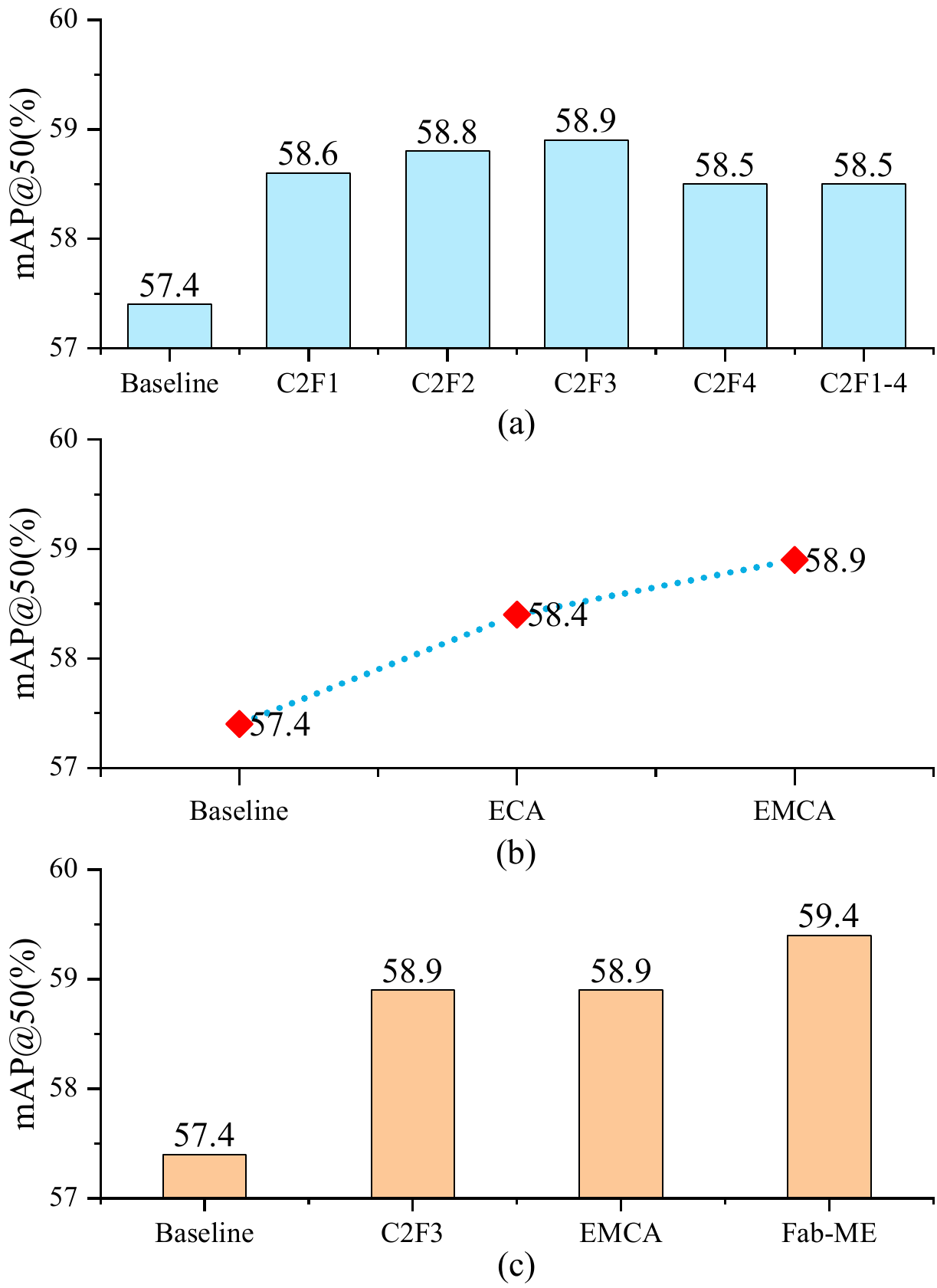}
\caption{Ablation study of key components in our method. (a) Ablation experiments of substituting the C2F modules at different positions within the Neck with the C2F-VMamba module. In the neck module illustrated in Fig. \ref{fig:framework}, the C2F modules are sequentially labeled from top to bottom as C2F1 through C2F4. Replacing each C2F module with the C2F-VMamba module results in models (2) through (5). (b) Ablation experiments of EMCA. (c)  Ablation study of key components in our method.}
\label{fig:ablation_image}
\vspace{-0.40cm}
\end{figure}

\begin{figure}[t]
    \centering
    \includegraphics[width=1\linewidth]{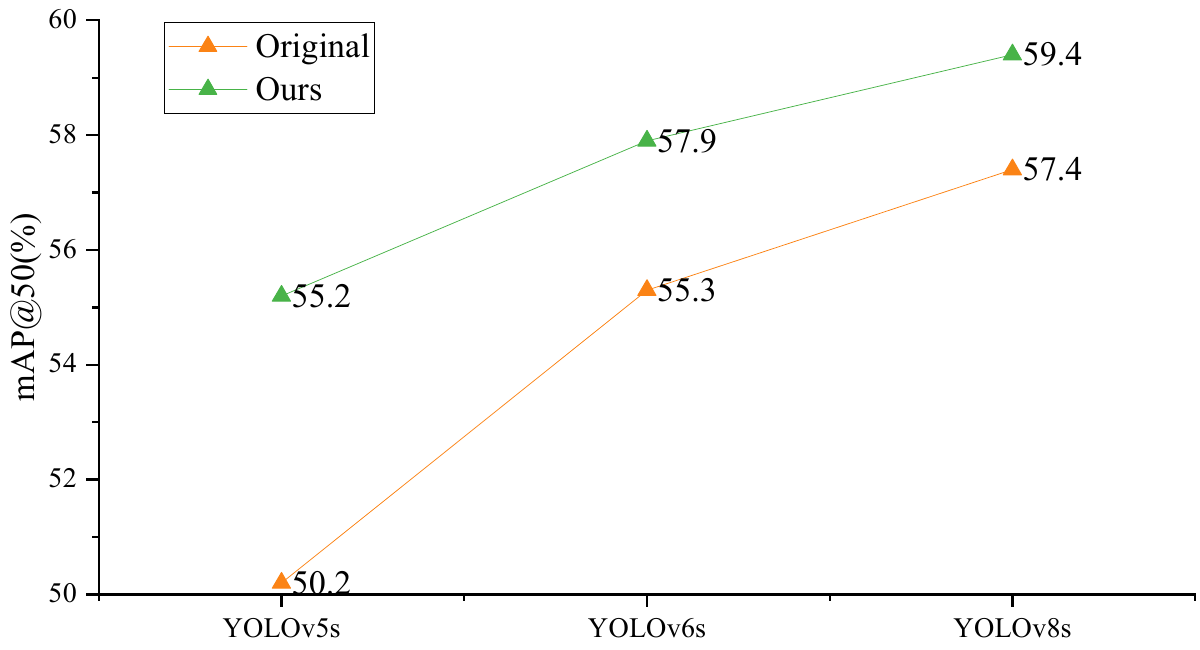}
    \vspace{-.2cm}
\caption{Ablation studies of our proposed module were conducted on YOLOv5s, YOLOv6s, and YOLOv8s.}
\vspace{-.2cm}
\label{fig:ablation_image_2}
\vspace{-0.50cm}
\end{figure}

\begin{figure*}[t]
    \vspace{-0.20cm}
    \centering
    \includegraphics[width=0.9\linewidth]{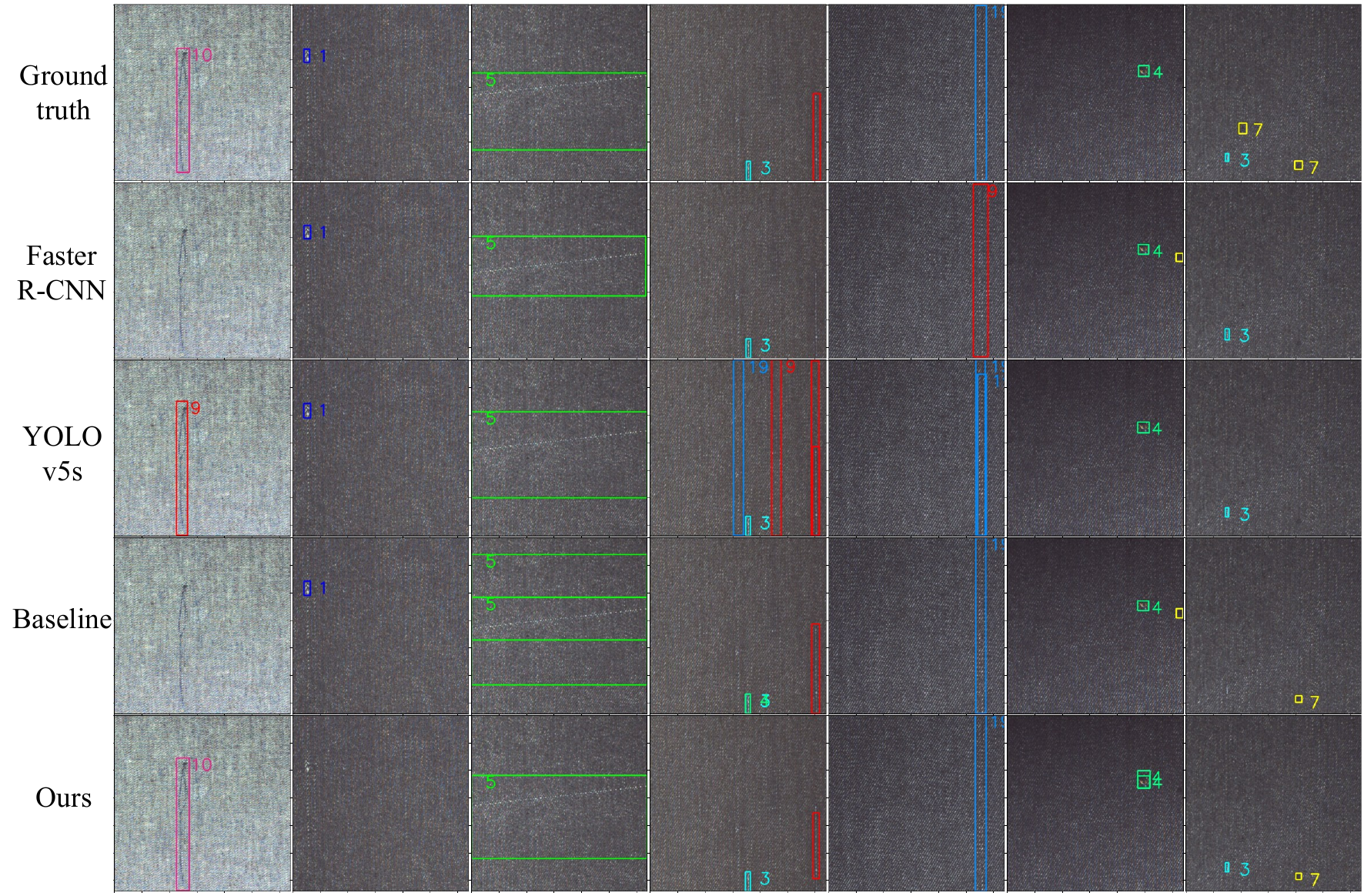}
\caption{Visualization result. Compared our method with Faster R-CNN, YOLOv5s, and baseline. The numbers near the boxes in the figure represent the category numbers of the defects, and the category names and colors can be found in the Name and Color column of Table~\ref{table:samples}.}
\vspace{-.2cm}
\label{fig:all_model}
\vspace{-0.20cm}
\end{figure*}

Table ~\ref{table:all_model_table} provides a comprehensive performance comparison of 11 fabric defect detection algorithms. Each algorithm demonstrates distinct attributes in terms of detection accuracy and processing speed. The algorithm in this study achieves mAP@50 of 59.4\%, significantly surpassing other methods. The proposed algorithm exhibits superior accuracy and reliability in detecting and localizing fabric defects.
Although comparatively lower, the mAP@0.5 scores for algorithms like YOLO-TTD, YOLOX-CATD, AFAM, PAMF, YOLOv5s, and YOLOv8s consistently exceed 0.5, indicating their practical applicability in defect detection. The algorithm presented in this study achieves high detection accuracy for fabric defect detection, offering robust technical support for practical applications.

\subsection{Ablation Studies and Analysis} 
The comparison results presented in Fig. \ref{fig:ablation_image}(c) demonstrate that the proposed Fab-ME method is superior to many state-of-the-art fabric defect detection methods. As shown in Fig. \ref{fig:ablation_image_2}, the C2F-VMamba and EMCA modules demonstrate their versatility across other YOLO versions, significantly enhancing the defect detection capabilities of the original YOLOv5s and YOLOv6s models.

In what follows, the proposed Fab-ME method is comprehensively analyzed from four aspects to investigate the logic behind its superiority.

\subsubsection{\textbf{Role of C2F-VMamba Module}}  
The C2F-VMamba module incorporates VSS blocks into the C2F block of the YOLOv8s feature fusion network. This integration significantly enhances the model's capability to capture intricate fabric defect details and broad contextual information. As shown in Fig. \ref{fig:ablation_image}(a), the global receptive field and linear complexity of the model significantly enhance detection accuracy. The most effective modification was replacing the third C2F in the Neck with C2F-Vmamba, resulting in a performance improvement of 2.5\%.
  
\subsubsection{\textbf{Influence of EMCA Attention Module}} 
The EMCA attention module, integrated into the final layer of the feature extraction network, substantially improves the model's multi-scale feature capture capabilities and sensitivity to detecting small targets. Our approach utilizes a dual-pooling strategy that integrates adaptive average pooling with adaptive maximum pooling. It encodes multi-scale contextual information, thereby enhancing detection performance.

\subsubsection{\textbf{Impact on Model Accuracy}}
The integration of the C2F-VMamba and EMCA modules substantially enhances model accuracy. As illustrated in Fig. \ref{fig:ablation_image}(c), the mAP@0.5 metric demonstrates a progressive increase with the sequential integration of the modules, culminating in a peak performance of 59.4\% when both modules are incorporated into the Fab-ME model.

\subsubsection{\textbf{Complexity Analysis}}
Despite the added layers and operations from the C2F-VMamba and EMCA modules, the Fab-ME model maintains a manageable complexity due to its lightweight design and efficient implementation, which enables the model to sustain high processing speeds, rendering it suitable for real-time applications.

\subsection{Visualization} 
The comparison diagram in Fig. \ref{fig:sample_image} presents the predictions of four algorithms on eight randomly selected images from the validation set. The detection results demonstrate that Faster R-CNN exhibits suboptimal performance, with frequently missed detections and false positives. The proposed model achieves superior detection performance, closely matching the original annotations. Specifically, the proposed algorithm successfully identifies the 10th defect type in the first column and the 3rd and 7th defect types in the seventh column, whereas the other three algorithms fail to do so effectively. The proposed algorithm demonstrates superior performance compared to the original YOLOv8s, consistent with prior comparative analyses.

\section{Conclusion}\label{sec:con}
We proposed Fab-ME, a real-time fabric defect detection framework based on YOLOv8s, designed to improve accuracy, efficiency, and global feature extraction. Fab-ME integrates the C2F-VMamba module, leveraging VSS blocks for enhanced global context and detail capture, and the EMCA module, which improves multi-scale sensitivity and small target detection.
Experiments on the Tianchi dataset showed a 3.5\% mAP@0.5 improvement over YOLOv8s, demonstrating the effectiveness of our approach. Future work will address challenges with subtle defects and explore advanced attention mechanisms and diverse datasets to further enhance robustness and detection performance.

\fontsize{8pt}{9pt}\selectfont
\bibliographystyle{IEEEbib}
\bibliography{ref}
\end{document}